\documentclass[letterpaper, 10 pt, conference]{ieeeconf}  

\usepackage{amsmath,amsfonts}
\usepackage{algorithmic}
\usepackage{algorithm}
\usepackage{array}
\usepackage[caption=false,font=normalsize,labelfont=sf,textfont=sf]{subfig}
\usepackage[capitalise]{cleveref}
\crefname{figure}{Fig.}{Figs.}
\Crefname{figure}{Fig.}{Figs.}
\crefname{section}{Sec.}{Secs.}
\Crefname{section}{Sec.}{Secs.}
\usepackage{textcomp}
\usepackage{stfloats}
\usepackage{url}
\usepackage{verbatim}
\usepackage{graphicx}
\usepackage{cite}
\usepackage{xcolor}
\IEEEoverridecommandlockouts                              

\title{\LARGE \bf
Actuation space reduction to facilitate insightful shape matching in a novel reconfigurable tendon driven continuum manipulator
}
\author{Sabyasachi Dash$^{1}$, John Golden$^{2}$ and Girish Krishnan$^{1\dagger}$
\thanks{This material is based upon work supported in part by the Air Force Office of Scientific Research (AFOSR), as part of the Space University Research Initiative (SURI), under award number FA9550-23-1-0723, and NSF-USDA COALESCE (Grant Number: 2021-67021-34418)}
\thanks{$^{1}$Sabyasachi Dash and Girish Krishnan are with the Department of Industrial and Enterprise Systems Engineering, University of Illinois Urbana-Champaign. {$^{2}$}John Golden is with the Department of Mechanical Science and Engineering, University of Illinois, Urbana-Champaign.$^{\dagger}$Corresponding author{ \tt\small gkrishna@illinois.edu} }%
}

\begin{document}

\maketitle
\thispagestyle{empty}
\pagestyle{empty}

\maketitle

\begin{abstract}
In tendon driven continuum manipulators (TDCMs), reconfiguring the tendon routing enables tailored spatial deformation of the backbone. This work presents a design in which tendons can be rerouted either prior to or after actuation by actively rotating the individual spacer disks. Each disk rotation thus adds a degree of freedom to the actuation space, complicating the mapping from a desired backbone curve to the corresponding actuator inputs. However, when the backbone shape is projected into an intermediate space defined by curvature and torsion (C-T), patterns emerge that highlight which disks are most influential in achieving a global shape. This insight enables a simplified, sequential shape-matching strategy: first, the proximal and intermediate disks are rotated to approximate the global shape; then, the distal disks are adjusted to fine-tune the end-effector position with minimal impact on the overall shape. The proposed actuation framework offers a model-free alternative to conventional control approaches, bypassing the complexities of modeling reconfigurable TDCMs.
\end{abstract}

\section{Introduction}
\begin{figure}[h!]
    \centering
    \includegraphics[width=0.47\textwidth]{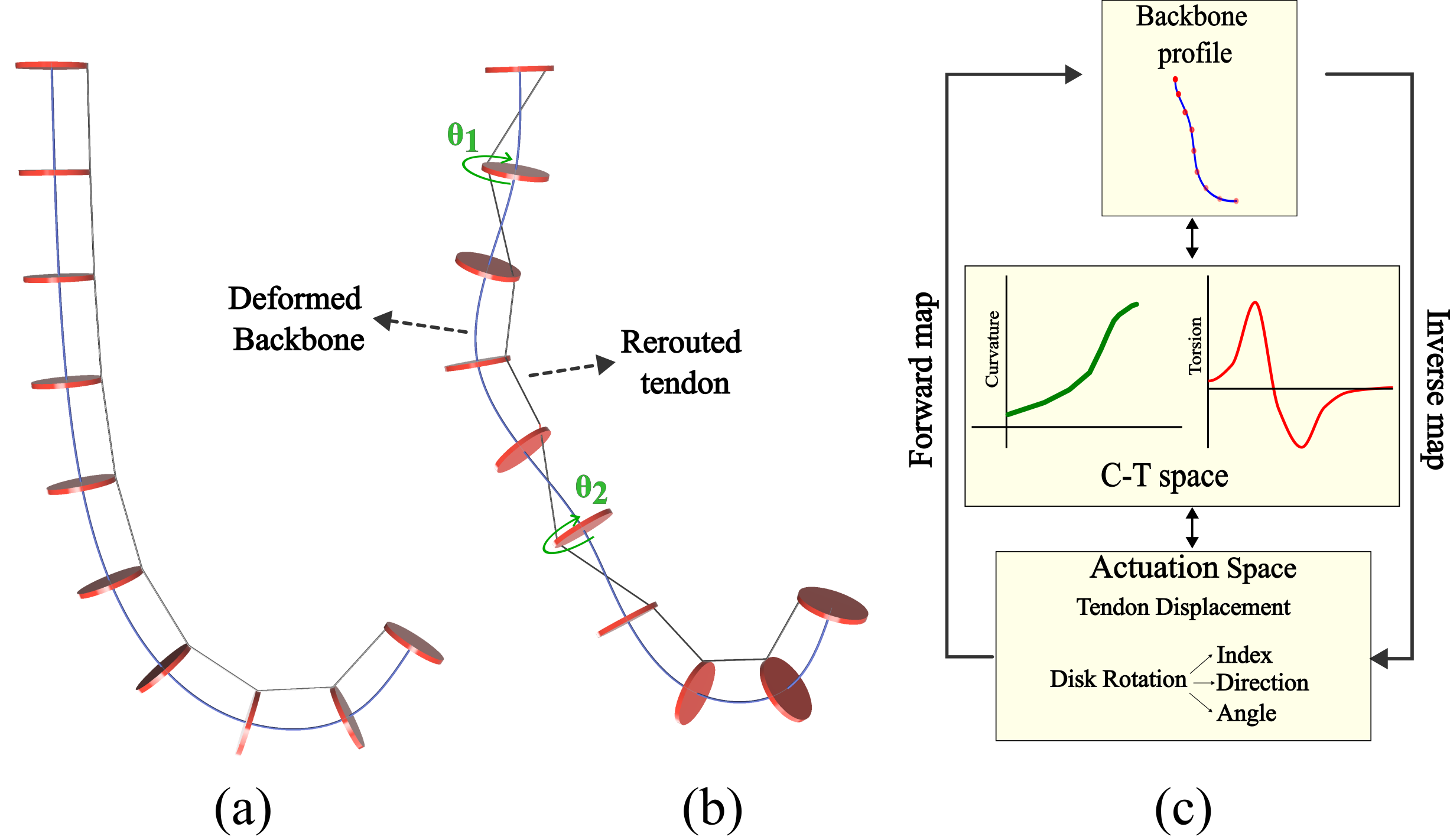} 
    \caption{Conventional TDCMs (a) rely on fixed tendon routing which limits the range of achievable spatial defomration modes. The proposed Reconfigurable TDCM (RTDCM) design (b) enables local tendon rerouting via controlled rotation of intermediate spacer disks, allowing more versatile backbone spatial configurations. (c) The geometric Curvature-Torsion (C-T) space provides an intermediate mapping between the manipulator's spatial deformation and the corresponding actuation space.}
    \label{design_domain_venn}
\end{figure}

Tendon-driven continuum manipulators (TDCMs) \cite{Walker2013} represent a prominent class of soft robotic systems, characterized by their ability to generate smooth, continuous deformations through tendons routed along a flexible backbone. Their intrinsic compliance, dexterity, and capacity to maneuver within highly constrained environments have made them well‑suited for applications such as minimally invasive surgery, search and rescue, and industrial automation \cite{burgner2015continuum}. The deformation of a TDCM is governed by (a) the tensions applied to the tendons and (b) the routing direction and orientation of those tendons. While tendon tensions directly influence the magnitude of deformation, the routing configuration dictates the set of deformation modes. These modes are fundamentally constrained by the number of tendons and the fixed geometric routing embedded in conventional designs. For instance, the most common routing scheme employs tendons running parallel to the backbone, producing bending toward the activated tendon. Variations in tendon routing, including helical and polynomial patterns \cite{8206554,5957337}, have been explored, yet the range of achievable deformation modes remains fundamentally limited by the fixed nature of these routing schemes. A more modular design was proposed in \cite{grassmann2022fas} where the extrinsic rotation of the backbone enabled twist in the entire manipulator. However, the intermediate spacer disks were floating and rotated passively rather than being actively actuated. In this paper, we introduce an alternate design framework to enhance the deformation modes through reconfiguring tendon routing by enabling independent rotation of each spacer disk. In our approach, the spacer disks traditionally used to guide the tendons are capable of rotating about the backbone axis (\cref{design_domain_venn}(b), \cref{fig:design_modular}) enabling dynamic modulation of tendon paths during operation. The reconfigurable tendon-driven continuum manipulator (RTDCM) can generate a broader repertoire of deformation behaviors, including unidirectional bending, bi-directional bending, and other fixed and arbitrary shape profiles, by controlling the rotational degree of freedom for each intermediate disk. We envision the use of these highly dexterous manipulators in agricultural manipulation for intricate plant imaging and in space applications for satellite repair.

In soft robotic manipulators, inverse kinematics involves determining an actuation strategy that produces a desired end-effector target or a backbone shape\cite{DellaSantina2023}. This mapping between the actuation space and task space is typically facilitated by a model, which can be inverted analytically or computationally for shape matching. RTDCMs introduce additional complexity in inverse kinematics computation: they feature a high dimensional actuation space that includes tendon tensions and disk rotations, and a nonlinear mapping between these actuation inputs and deformed shapes, precluding insights. This paper first seeks to develop insight into the relationship between the actuation space and the resulting spatial deformations by projecting the RTDCM’s shape into a curvature–torsion (C–T) space \cite{kuhnel2005differential} (\cref{design_domain_venn}(c)). Similar projection‑based approaches such as representations in strain space \cite{Kim2022} and tendon‑displacement space \cite{Grassman2025} have been proposed to simplify this mapping. For simpler continuum manipulators, constant‑curvature models \cite{neppalli2009,websteriii2010,DellaSantina2020} and rigid‑body approximations \cite{zhang2018, kolpashchikov2022,wu2022,zhang2023} have proven effective. We show that the C-T space inherently encodes qualitative information about tendon routing patterns induced by specific spacer disk rotations. In particular, the torsion profile of a spatial curve reveals the indices and directions of disk rotations. Our analysis identifies two maximally decoupled disk sets for an RTDCM: (1) the proximal and intermediate disks, which govern global shape deformation, and (2) the distal disks, which control end-effector positioning. These insights enable a principled reduction of the dimensionality of the actuation space. Shape matching can now be cast as a four-stage sequential approach: Step 1: Identifying the indices and direction of disk rotation from the Torsion information, Step 2: Identifying the tendon actuation that minimizes error from the desired curvature, Step 3: Obtaining the disk rotation angles that minimize the global shape error, and Step 4: Fine-tuning the end effector to meet target tip position. 

The rest of the paper is organized as follows: \cref{design} demonstrates the RTDCM design and the experimental framework, followed by \cref{analysis} which explores empirical insights from the C-T space. \cref{control} presents the framework for the four-stage sequential actuation, and  
\cref{results} illustrates two shape matching examples using the framework.



\section{Design and Experimental Methods}
\label{design}
\subsection{Manipulator Design for tendon rerouting}
\begin{figure}[h]
   \centering    \includegraphics[width=0.32\textwidth]{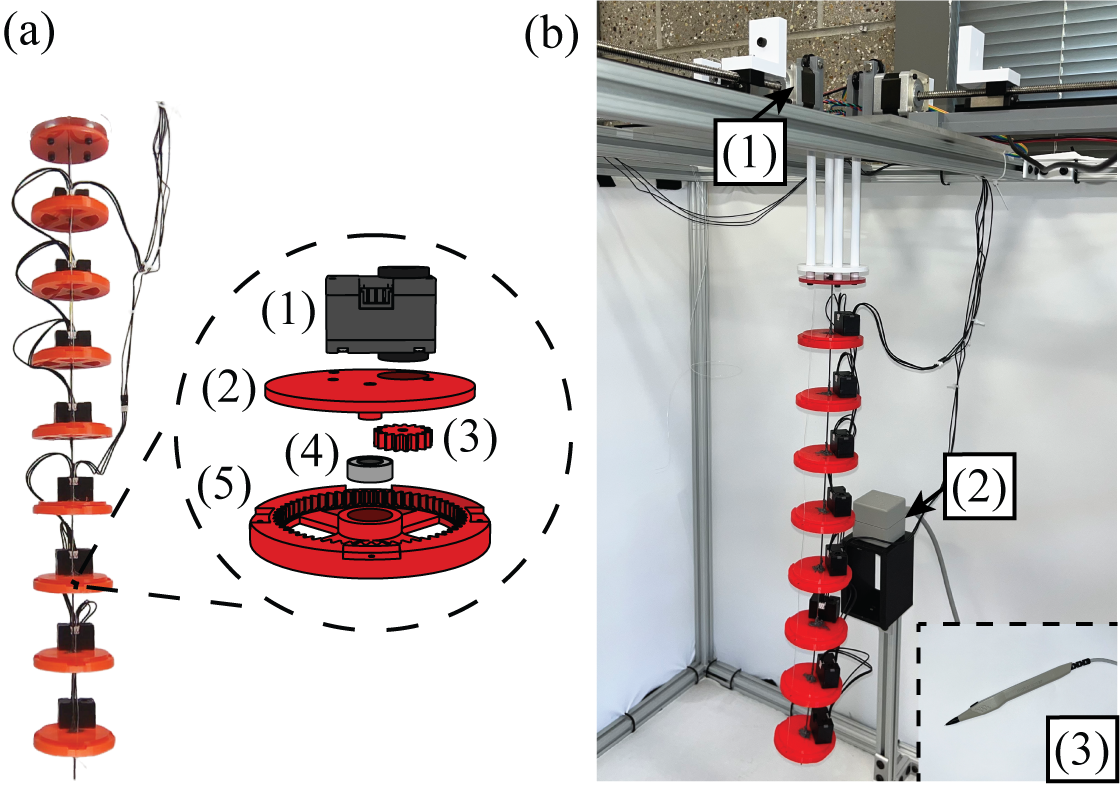} 
   \caption{(a) RTDCM design with each disk assembly capable of rotating the tendon locally. The servo motor (1) sits on the fixed mount (2) which is attached to the backbone using epoxy. The fixed mount is connected to the rotating disk (5) through a bearing (4), and a gear (3) attached to the servo motor-head rotates (5) which holds the tendon. (b) Experimental setup with stepper motor (1) to actuate the tendon, and a magnetic stylus (3) with reference at (2) to retrieve track coordinates of the backbone.}
   \label{fig:design_modular}
\end{figure}
We implement a novel design to locally rotate the spacer disks using servo motors, thereby enabling tendon rerouting.
As demonstrated in \cref{fig:design_modular}(a), each spacer disk features a fixed mount (2) which is rigidly attached to a 1.5 mm thick Nitinol backbone using steel-filled epoxy, and holds the servo motor (1). The servo motor DYNAMIXEL XL330-M288-T (ROBOTIS Co. LTD, Gangseo-gu, Seoul, Korea), with an absolute multi-turn encoder helps track the angular position of the tendon. All motors are connected under a shared communication bus using an OpenRB-150 embedded controller. The fixed mount (2) is connected to the rotating disk (5) concentrically via a bearing (4) and a spur gear (3) meshed between them. Upon actuation of the servo, the rotating disk (5) alters the tendon’s angle, thereby locally rerouting the tendon within the segment between the disks above and below it. This automatic disk rotation enables tendon rerouting both prior-to and after tendon actuation; the latter is leveraged in the shape matching examples presented in this paper. The manipulator considered is 560 mm long with eight equal length segments, and the tendon passes through a hole on each disk at a radial distance of 34 mm from the backbone centerline.
\subsection{Experimental Setup}
\label{setup}
The manipulator is suspended under gravity inside a closed frame. Horizontally mounted NEMA-17 stepper motors actuate the tendons. Though four such motors can displace the tendons in mutually perpendicular directions, we consider only a single tendon for this paper, leveraging the disk rotations to navigate through the spatial workspace. We restrict the servos on each disk to rotate $90^\circ$ in both clockwise and anticlockwise directions and tendon displacements to up to 140 mm. A 6-DOF magnetic stylus is manually pointed to the center of each disk to reconstruct the 3D curve after deformation. The entire experimental setup is shown in \cref{fig:design_modular}(b).
To account for any errors in measurement, repeated data was collected by pointing to the disk centers multiple times, and a density based clustering, DBSCAN \cite{10.5555/3001460.3001507} was implemented to find the centroid of each data cluster.

\section{Analysis}
\label{analysis}
\begin{figure*}[!t]
    \vspace{5pt}
    \centering    \includegraphics[width=0.72\textwidth]{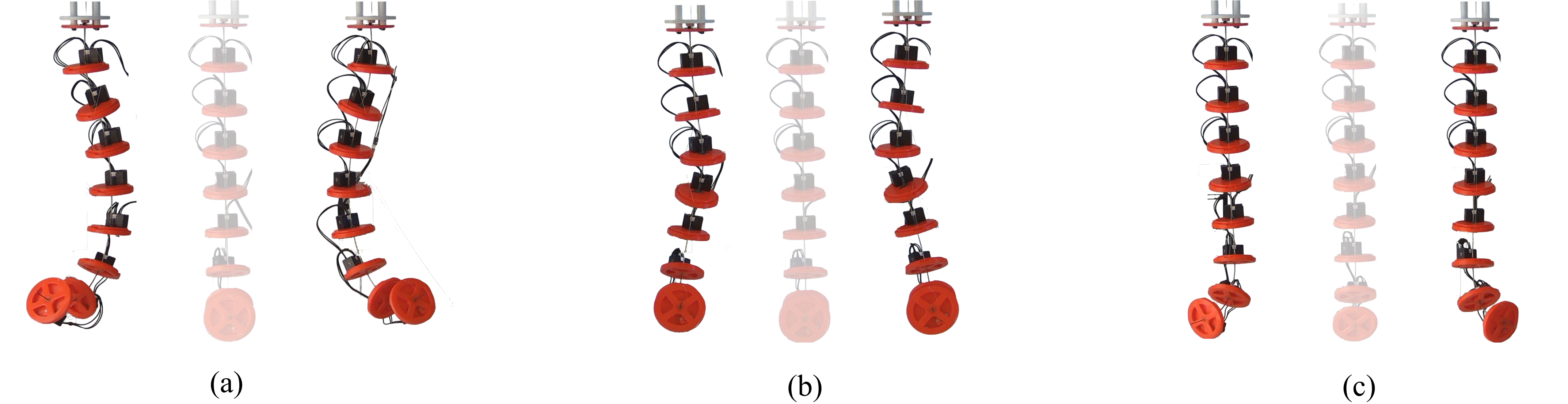} 
    \caption{Continuous rotation of (a) second, (b) fourth, (c) eighth disk from $90^{\circ}$ clockwise to $90^{\circ}$  anticlockwise direction, where (a) proximal actuation yields behavior similar to entire base rotation, (b) intermediate actuation induces a more pronounced global shape change of the backbone, and (c) distal actuation enables local tip maneuvering with minimal effect on the overall shape}
    \label{rotation}
\end{figure*}
\label{c-tspace}
This section discusses various empirical inferences from the manipulator's deformation behavior, which lays the framework for the sequential actuation guidelines for shape matching, as will be presented in \cref{control}. The inferences are backed by qualitative mechanics-based arguments, and a strict analytical proof is beyond the scope of this paper. 
\subsection{Rotated disk index affects the manipulator's behavior}
The manipulator's behavior is strongly influenced by the index of the rotated disk. Consider \cref{rotation} where a single disk: second (proximal), fourth (intermediate) and eighth (distal), is rotated continuously sweeping through $90^\circ$ in clockwise and counterclockwise directions. For each case, a single tendon is pulled to a fixed displacement. Rotating the proximal disk \cref{rotation}(a) creates a change in curvature only near the top, while the rest of the manipulator remains largely in plane, demonstrating an effect similar to rotating the entire base which can be useful in applications requiring a broad sweep of the workspace. Rotating an intermediate disk largely influences global shape change, without affecting the end-effector's pose significantly, as demonstrated in \cref{rotation}(b). This is important for shape-matching applications, where identifying the intermediate disk(s) responsible for the desired shape change can significantly reduce the actuation space complexity. Finally, any adjustment to the end effector can be achieved by rotating the distal disks. For this paper, we consider the end-effector to be controlled by the eighth (penultimate distal) disk, thus affecting the disk above and below it. After achieving an approximate global shape through intermediate disk actuation, the distal disk can be rotated to reach the target tip pose.
\subsection{Curvature-Torsion space Analysis}
As discussed previously, mapping the target spatial curve to the actuation space is inherently complicated due to the large number of actuation parameters (in our case: eight disks, each of which can rotate in clockwise or counterclockwise direction, and the tendon). In order to reduce the dimensionality, we leverage insights from an intermediate Curvature-Torsion (C-T) space (\cref{design_domain_venn}(c)), which inspire the sequential actuation framework.
\subsubsection{Curvature-Torsion Space Projection}
Spatial deformation of a curve can be characterized by a combination of curvature and torsion using the Frenet-Serret formulas given by
\begin{equation}
\label{equationcurve}
\kappa = \frac{\left\| \mathbf{r}'(s) \times \mathbf{r}''(s) \right\|}{\left\| \mathbf{r}'(s) \right\|^3},
\quad
\tau = \frac{ \left( \mathbf{r}'(s) \times \mathbf{r}''(s) \right) \cdot \mathbf{r}'''(s) }{ \left\| \mathbf{r}'(s) \times \mathbf{r}''(s) \right\|^2 }
\end{equation}

where $\mathbf{r}$ is the position vector of the deformed curve, and $\mathbf{r}'(s)$, $\mathbf{r}''(s)$, and $\mathbf{r}'''(s)$ are the first, second, and third derivatives of the position vector with respect to the manipulator arc length. 
The curvature is the intensity at which the spatial curve bends in plane, signified by the rate of change of the tangent vector. Torsion, on the other hand, measures the change in the osculating plane, giving information about any out-of-plane deformation. Variations in the C-T space can thus yield important insights about the curve's deformation. For all our experiments, we use the disk center coordinates measured using the magnetic stylus as discussed in \cref{setup}. Since \cref{equationcurve} relies on the second and third derivatives of the position vector, the C-T behavior is prone to artificial noise and spikes. Therefore, care was taken to not excessively smoothen the curve from the measured position data. Instead, a spline-based interpolation was implemented to the C-T profiles after computing them from \cref{equationcurve} in order to achieve a smooth representation.
\vspace{-1pt}
\begin{figure*}[!t]
    \vspace{5pt}
    \centering        \includegraphics[width=0.81\textwidth]{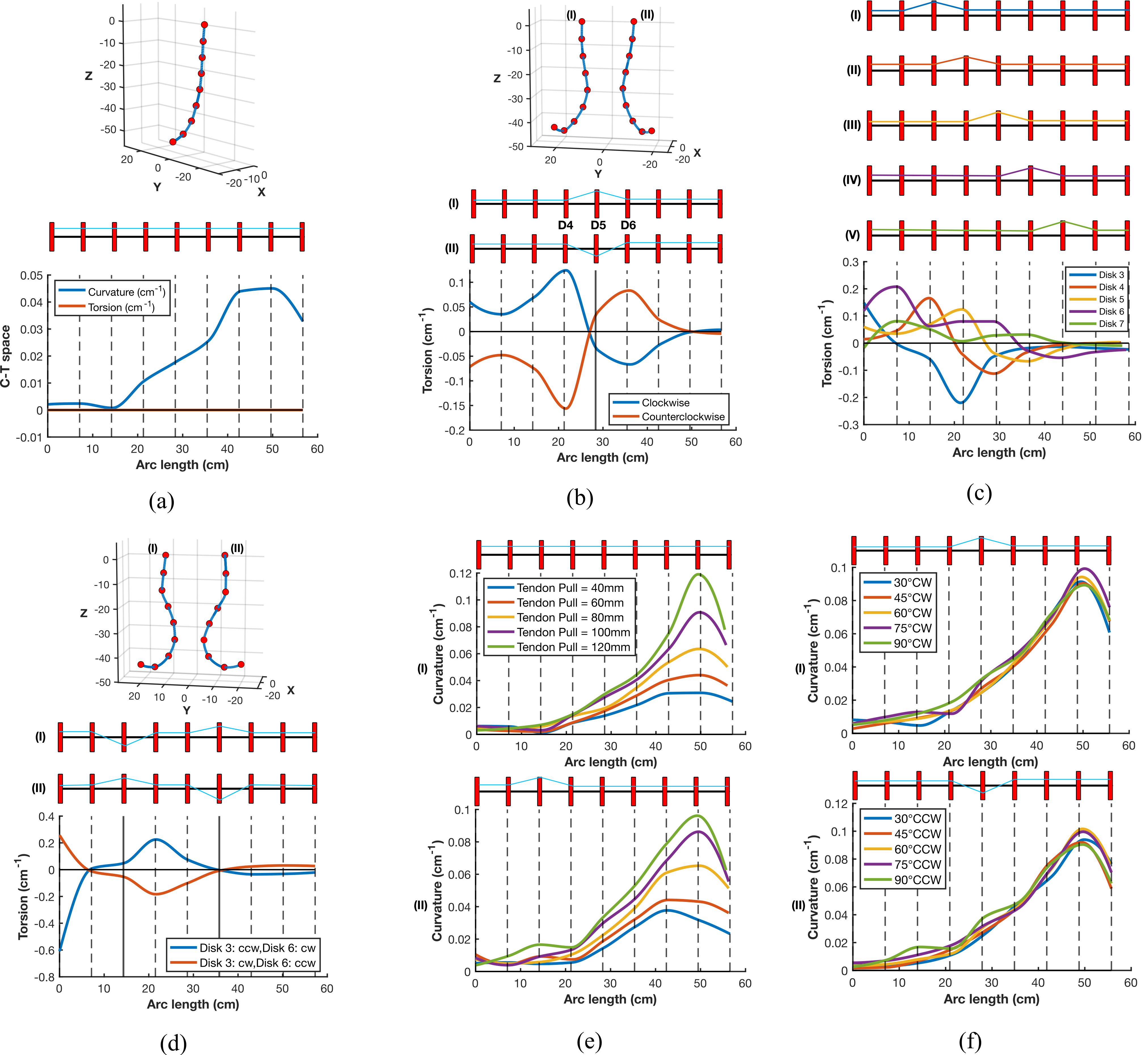} 
    \caption{(a) Parallel tendon routing yields a planar deformation profile of the backbone, which is signified by zero torsion throughout, and a rising curvature due to gravity. (b) Spatial deformation induced by local rotation of the fifth disk in (I) clockwise and (II) counterclockwise direction. Torsion plot demonstrates a sign change near the rotated disk, with the signs flipped between the clockwise and anticlockwise case. (c) Torsion plots for $90^\circ$ clockwise rotation of (I) third through (V) seventh disk. The torsion flips sign from positive to negative in close proximity of each rotated disk. (d) Torsion plots for combination of two disks (third and sixth) rotated in opposite directions demonstrate two distinct sign changes. (e) Increasing tendon displacement directly correlates with a steeper curvature for both (I) parallel routing, and (II) routing with a rotated disk, while the peak curvature and slope reduce for the latter case. (f) For a fixed tendon displacement, the fifth disk is rotated incrementally from $30^\circ$ to $90^\circ$, both in (I) clockwise and (II) anticlockwise directions. The resulting curvature profiles remain close to each other in both cases.}
    \label{strainspace}
\end{figure*}
\subsection{Inferences from C-T space}
\label{inferences}
\subsubsection{No torsion is induced when the tendon is routed parallel to the backbone}
First, a baseline case is demonstrated in \cref{strainspace}(a) where a tendon is routed parallel to the manipulator's backbone and none of the disks are rotated, emulating conventional TDCMs. As expected, there is no significant change in torsion, pertaining to the pure in-plane bending of the manipulator. The curvature rises due to the effects of gravity, which would ideally be constant for a manipulator having negligible weight.  
\subsubsection{Rotating a single disk induces a local sign change in Torsion}
Consider \cref{strainspace}(b), where only the fifth disk is rotated clockwise. This isolated rotation causes a localized clockwise shift in the tendon segment between the fourth (D4) and fifth (D5) disks, as seen in the upper part of \cref{strainspace}(b). The tendon’s reorientation leads to a change in its curvature and introduces a slight mechanical twist in the backbone. These effects collectively result in a pronounced torsional response in the C–T space, evident as the left peak in the torsion curve. Immediately following this segment, between the fifth and sixth disks, the tendon begins to revert to its original orientation. This reversal causes the torsion to change sign and eventually approach zero, as observed in \cref{strainspace}(b). Similar torsion sign changes were observed when rotating disks $i = 3, 4, 6 , 7$, as shown in \cref{strainspace}(c). Ideally, the sign change would occur exactly where the disk is rotated. However, gravitational effects can shift this point. This is most clearly seen in the case of the third disk, where the torsion sign change occurs earlier than expected.
Moreover, the sign of torsion conveys its directionality. For instance, rotating the same disk counterclockwise (shown as orange in \cref{strainspace}(b)) produces a similar local sign inversion at the actuated disk. However, the signs of the leading and trailing segments are reversed compared to the clockwise case. In the manipulator configuration considered for this paper, a disk rotated in clockwise direction corresponds to a sign change from positive to negative and vice versa. Therefore, the torsion plot alone gives information about both the index, and direction of rotation of the actuated disk.
\subsubsection{Rotating two disks in opposite directions corresponds to two sign changes in torsion}
Here, we extend Inference 2 to scenarios involving the rotation of two or more disks. The torsion space can reveal actuation patterns when multiple disks are rotated in opposite directions. For example, in the torsion plot shown in \cref{strainspace}(d), two distinct sign changes appear near the second and sixth disks. In the real experiment, the third and sixth disks are actuated in opposite directions, dividing the backbone into three segments with alternating torsional directionality. As discussed previously, sharp shape changes are primarily driven by intermediate disks, which are effectively captured in the torsion profile. Additionally, torsion induced by gravity can either amplify or diminish the influence of one disk relative to another, shifting the torsion sign changes to nearby disks.
\vspace{-2pt}
\subsubsection{Curvature profiles correlate with applied tendon displacement}
As observed in \cref{strainspace}(a), gravitational effects yield a gradually rising curvature profile. Intuitively, increasing the magnitude of tendon displacement results in an increase in the curvature steepness, and hence the overall peak, as observed for the parallel manipulator in \cref{strainspace}(e)-(I). A similar behavior is observed in \cref{strainspace}(e)-(II), where one of the disks is rotated, but the maximum peak is relatively lower since the rotated disk enforces a drop in the magnitude of curvature. An important empirical observation can be made from \cref{strainspace}(f), where for a fixed tendon displacement and a specific disk rotated, the curvature profiles remain fairly unaffected with the magnitude of disk rotation. Consider the fifth disk rotated clockwise and counterclockwise in \cref{strainspace}(f)-(I) and (II) respectively. The peak-to-peak error observed in the curvature while the concerned disk is rotated from $30^\circ$ to $90^\circ$ is just 0.009 $cm^{-1}$ for each case. This can be leveraged to approximate the required tendon actuation for meeting a desired global shape change, even when the exact disk rotation is not known. 

\section{Sequential Actuation Framework}
\label{control}
\begin{figure}[]
\vspace{5pt}
    \centering
\includegraphics[width=0.42\textwidth]{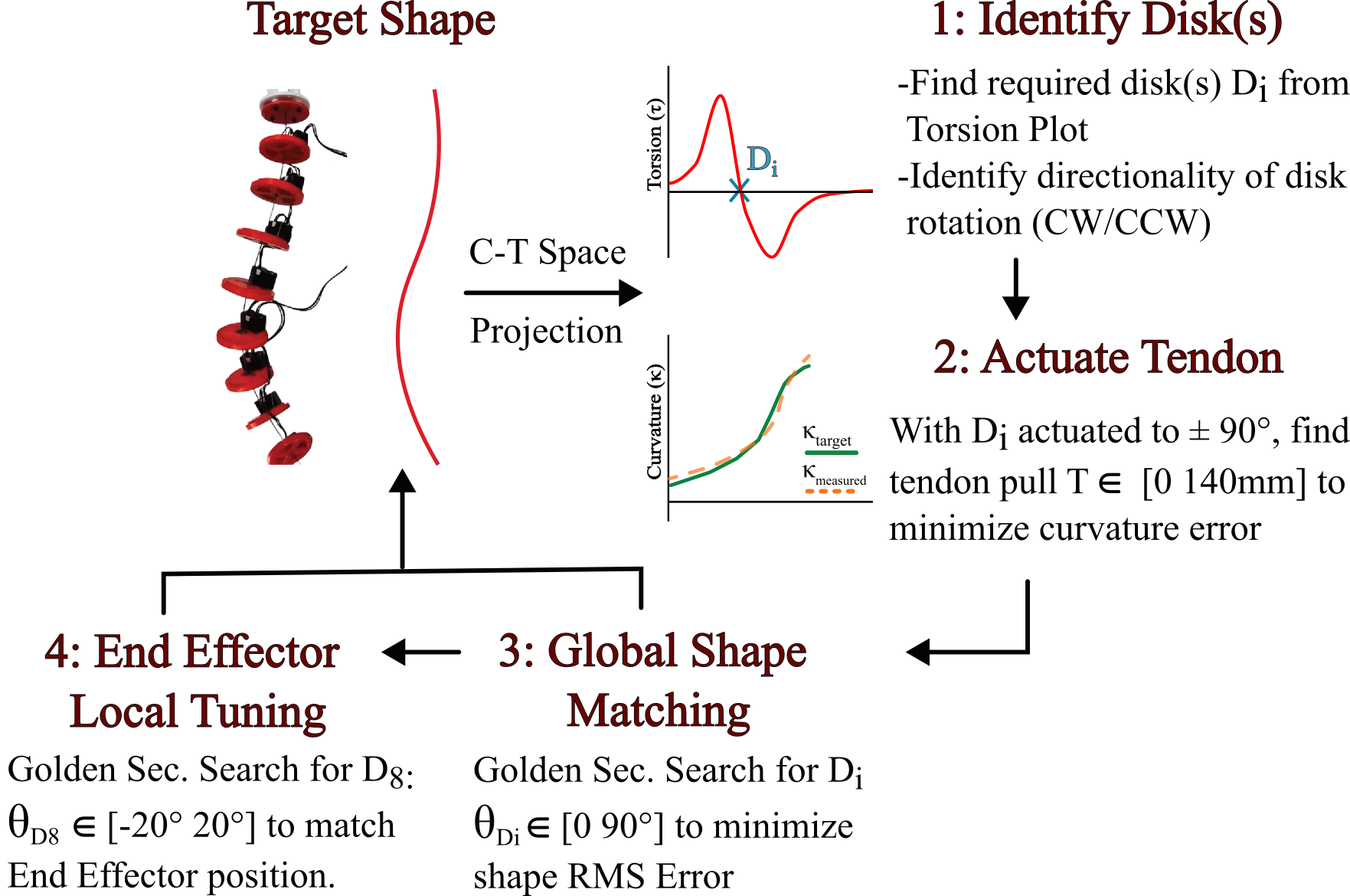} 
    \caption{Sequential Actuation Framework for Target Shape Matching}
    \label{controlalgo}
\end{figure}
The actuation space comprises of: (a) tendon displacement achieved by the stepper motor actuation, and (b) nine spacer disk rotation(s). Obtaining an inverse map for a desired backbone curve would thus involve: (i) the magnitude of tendon displacement, (ii) index of disks to be rotated, (iii) direction of rotation of each disk, and (iv) angle of disk rotation. Here, we demonstrate a sequential framework, which assists in simplifying the entire actuation space architecture, primarily using insights drawn from the Curvature-Torsion space, as discussed in \cref{c-tspace}.
The proposed sequential framework (\cref{controlalgo}) consists the following: First, the torsion plot gives information about the index, and directionality of rotation of the primary intermediate disk(s) responsible for global shape change. There may be more than two disks rotated in the same direction relative  to each other, which may not reflect as distinct sign reversals in the torsion plot. In such cases, the primary disks responsible for shape change can be first rotated to minimize the shape error, and any residual error can be adjusted by rotation of the end effector per Step 4. Once the disks are identified from Step 1, they are rotated to $90^\circ$ in their respective directions, and the tendon is pulled till the error from the desired curvature is minimized. Towards this, we use Root Mean Squared Error (RMSE) between target and attained curvatures, and a simple 1D optimization method - the Golden Section search to find the optimal solution. The Golden Section search works by iteratively narrowing the search interval using the golden ratio ($\approx 0.618$) to place the interior points. This effectively converges to a minimum without requiring any gradient computation. After obtaining the desired tendon actuation, Step 3 focuses on identifying the angle of rotation for each actuated disk. For the disk indices and directions identified from Step 1, a golden section search in the range of $[0\quad 90^\circ]$ is implemented to minimize the RMS Error of the resulting shape from the target. Finally, in Step 4, the end effector is actuated by rotating the eighth (penultimate distal) disk to guide the tip to its target position. Since actuating a disk affects its trailing and leading segments, we evaluate the RMS error in segments between the seventh and ninth (end) disks. To achieve this, we again implement a similar golden section search for the eighth disk between $20^{\circ}$ clockwise and $20^{\circ}$ counterclockwise directions. This sequential actuation process helps match the obtained shape with the target backbone profile. The next section will demonstrate implementation of this framework to two distinct shape-matching examples. 

\section{Results and Discussion}
\label{results}
\begin{figure*}[!t]
    \centering    \includegraphics[width=0.85\textwidth]{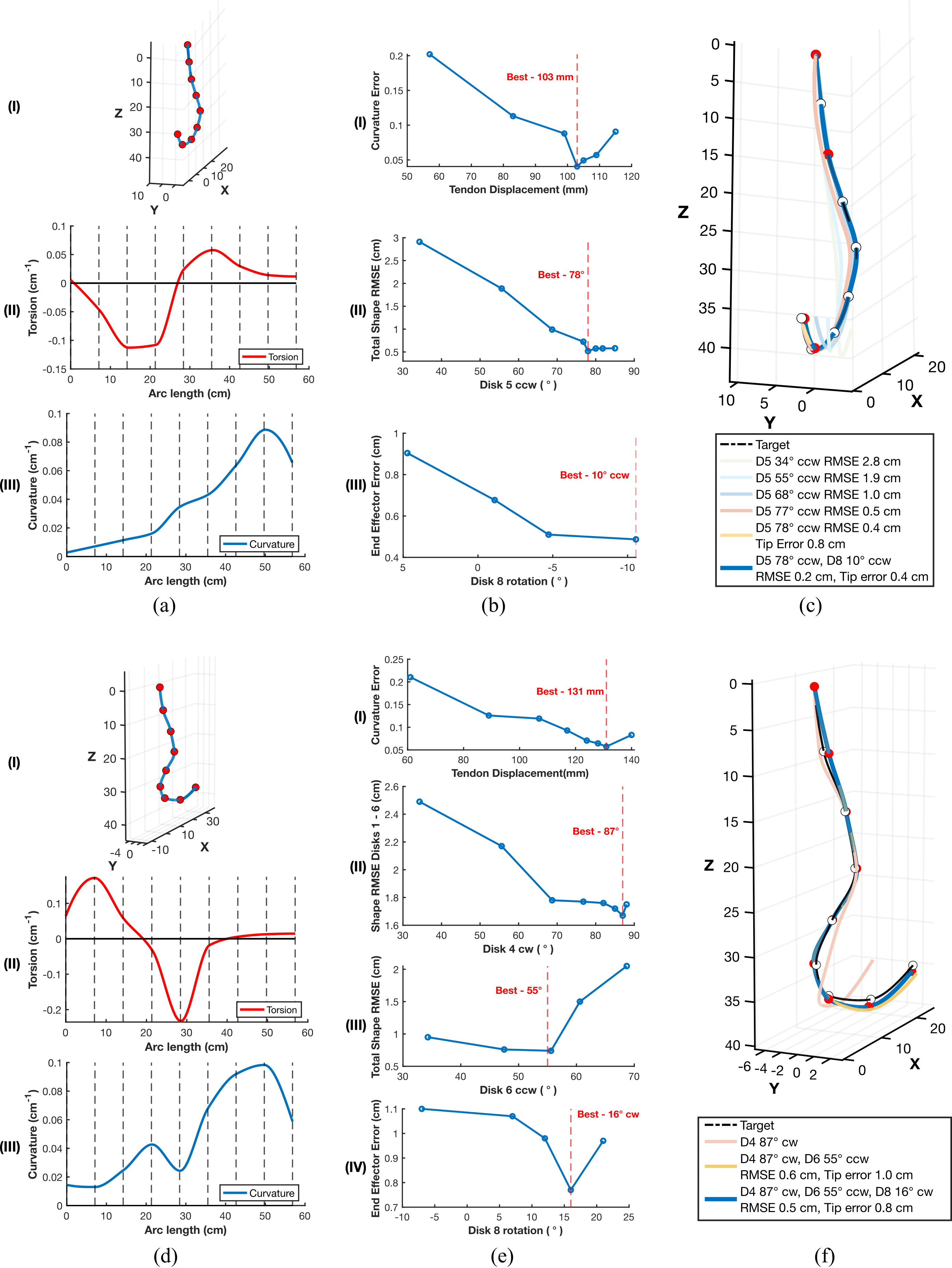} 
    \caption{(a) Target shape (I) with its corresponding torsion (II) profile demonstrating clear sign reversal from negative to positive near the fifth disk implying a counterclockwise rotation. (b) (I) With the fifth disk rotated to $90^\circ$, a tendon pull of 103 mm minimizes error from target curvature. (II) An exact angle of $78^\circ$ in the fifth disk minimizes the global shape error, and (III) a counterclockwise rotation of the eighth disk to $10^\circ$ minimizes the end-effector error. (c) Overlay of spatial curves obtained when Disk 5 is rotated iteratively till error with target shape is minimized, and Disk 8 is adjusted to match the tip.
    (d) A second target shape (I) with its torsion profile in (II), demonstrates a significant sign change near the fourth disk, and a minor sign change around the sixth/seventh disk. (e) (I) Both the fourth and sixth disks are rotated to $90^\circ$ clockwise and counterclockwise directions respectively, and a tendon displacement of 131 mm minimizes the curvature error. Required disks are rotated sequentially: (II) fourth disk first to $87^\circ$ clockwise, and (III) sixth disk next $55^\circ$ counterclockwise to minimize overall shape error. (IV) End effector is adjusted by rotating the eighth disk clockwise by $16^\circ$. (f) Overlay of Target shape with shapes obtained after rotation of fourth disk, fourth + sixth disk, and fourth + sixth + end effector.}
    \label{inverseexamples}
\end{figure*}
This section demonstrates two separate examples for obtaining the actuation space parameters to match desired backbone shapes, implementing the framework presented in \cref{control}. In all these examples, only a single tendon is pulled and the disks are constrained to rotate $90^\circ$ in each direction. The first target shape demonstrates a single sign reversal in the torsion plot, and the second shows two sign changes. 
\subsection{Target shape with a single torsion sign change:}
Consider the target shape in \cref{inverseexamples}(a)-(I), and its corresponding torsion plot in (II) signifying one prominent sign change from negative to positive near the fifth disk. As discussed in \cref{inferences}, this sign reversal corresponds to an actuation of the fifth disk in the counterclockwise direction. This completes Step 1 of the sequential framework, giving information about the index and directionality of the required disk(s) to be rotated. We proceed to Step 2 where the fifth disk is rotated to $90^{\circ}$ in the counterclockwise direction, and the tendon is pulled iteratively in the range from 0 to 140 mm with the search brackets narrowing down after every golden section iteration. This gives us an approximate tendon displacement of 103 mm (\cref{inverseexamples}(b)-I), and actuating the tendon above or below that value increases the curvature error. Next, we proceed to shape matching (Step 3), where we rotate the fifth disk in counterclockwise direction to $34.4^\circ$, $55.6^\circ$ and so on, based on the golden ratio split between $0^{\circ}$ and $90^{\circ}$. We only consider whole-numbered magnitudes of rotation pertaining to the precision of the servo motors. The minimum RMSE (0.4 cm) between the target and attained shape is obtained at $78^{\circ}$ rotation, as reported in \cref{inverseexamples}(b)-II. This configuration brings the manipulator in close proximity to the target shape, though the tip does not coincide entirely. From the torsion plot in \cref{inverseexamples}(a)-(II), the value of torsion does not fall completely to zero after the fifth disk, signifying that rotating some of the disks at the distal end may still be necessary. Therefore, we follow Step 4 to fine tune the end effector (by rotating the penultimate distal disk) to reduce the error of the tip. With a similar golden search between $[-20^{\circ},\ 20^\circ]$ (negative sign corresponding to counterclockwise rotation), we arrive at $\approx 10^\circ$ counterclockwise rotation as the best minimizer of the shape error for the distal segment. \cref{inverseexamples}(c) showcases the various iterations taken by the golden search algorithm which reaches significantly close to the target shape at $78^\circ$ disk 5 angle. Rotating the eighth disk (end-effector) in the counterclockwise direction slightly by $10^\circ$ further reduces the tip error to 4 mm.  
\subsection{Target shape with two torsion sign changes:}
Here, the target shape (\cref{inverseexamples}(d)-(I)) demonstrates a distinct torsion sign change (II) at the fourth disk from positive to negative signifying a clockwise rotation, and a minor sign reversal in between the sixth and seventh disk. Though rotation of the fourth disk is imminent, it is not clear if the sixth or the seventh disk needs to be actuated. As discussed earlier, the end effector actuated by the eighth disk affects the distal segment, so any error towards the end can be adjusted by the end effector fine tuning following Step 4. Therefore, we consider the fourth and sixth disks to be responsible for the shape matching. Proceeding to Step 2, we now rotate both the fourth and sixth disks to $90^\circ$ in clockwise and counterclockwise directions respectively, and evaluate the error from the target curvature by only actuating the tendon. A tendon displacement of 131 mm minimizes the error and is thus considered to be the required actuation. Per Step 3, it is necessary to determine the magnitude of rotation of the fourth and sixth disks. To consider their independent effect, the fourth disk alone is first considered for matching the target shape from the first till sixth disks, since the effect of the fourth disk can be realized in the segments above and below it. This yields a local minimizer at $87^\circ$, as shown in \cref{inverseexamples}(e)-(II). Next, we perform the 1D optimization for the sixth disk while the fourth disk has already been rotated, and now match the entire shape with the target. As demonstrated in \cref{inverseexamples}(e)-(III), a counterclockwise rotation of $55^\circ$ in addition to the fourth disk rotation of $87^\circ$ takes the manipulator reasonably close to the target shape, with an RMSE of 0.6 cm and tip error of around 10 mm. With further adjustments to the end effector per Step 4, the eighth disk is rotated to $16^\circ$ in the clockwise direction, to reduce the tip error slightly to 8 mm. 
\section{Conclusion and Future Work}
Tendon Driven Continuum Manipulators with reconfigurable tendon routing open avenues for achieving a large space of deformations. This work focuses on one such design, where each disk can rotate actively during operation enabling localized tendon rerouting. Although directly mapping manipulator deformations to the actuation space consisting of tendons and internal disks is non-trivial and fairly complicated, projecting these deformations into an intermediate Curvature-Torsion (C-T) space reveals valuable insights. Using these qualitative insights, we propose a sequential four-step actuation framework for reducing the degrees of freedom of the actuation inputs. Two examples are demonstrated that implement the proposed framework to obtain the actuation parameters for matching a desired spatial shape of the backbone. More complicated and dynamic shape changes will be explored in the future, incorporating real-time feedback from external cameras for backbone shape reconstruction. 

While the C-T space in this work provides important qualitative insights, future work will involve incorporating Cosserat rod mechanics to increase the fidelity and accuracy of the framework.
Furthermore, future design efforts will be directed towards reducing the weight of the manipulator and adding more intermediate modular disks for obtaining a wider range of deformations. Such a manipulator with a large number of independently controllable disks could substantially benefit from the proposed C-T based actuation space reduction framework.  

\bibliographystyle{IEEEtran}
\bibliography{references}
\end{document}